\documentclass[sigconf]{acmart}
\AtBeginDocument{%
  \providecommand\BibTeX{{%
    \normalfont B\kern-0.5em{\scshape i\kern-0.25em b}\kern-0.8em\TeX}}}

\copyrightyear{2024}
\acmYear{2024}
\setcopyright{acmlicensed}\acmConference[HRI '24 Companion]{Companion of the 2024 ACM/IEEE International Conference on Human-Robot Interaction}{March 11--14, 2024}{Boulder, CO, USA}

\acmDOI{XX.XXXX/XXXXXXX.XXXXXXX}

\acmConference[HRI 2024]{Human Robot Interaction 2024}{March 11--14,
  2024}{Boulder, Colorado, USA}
\acmBooktitle{Companion of the 2024 ACM/IEEE International Conference on Human-Robot Interaction (HRI '24 Companion), March 11--14, 2024, Boulder, CO, USA}
\acmISBN{XXX-X-XXXX-XXXX-X/XX/XX}

\usepackage{multirow}
\begin{document}

\title{DogSurf: Quadruped Robot Capable of GRU-based Surface Recognition for Blind Person Navigation}


\author{Artem Bazhenov}
\authornote{Both authors contributed equally to this research.}
\email{Artem.Bazhenov@skoltech.ru}
\orcid{0009-0001-2228-7298}
\affiliation{%
      \institution{Skolkovo Institute of Science and Technology}
      \city{Moscow} 
      \country{Russia}
}

\author{Vladimir Berman}
\authornotemark[1]
\email{Vladimir.Berman@skoltech.ru}
\orcid{0000-0003-4530-3518}
\affiliation{%
      \institution{Skolkovo Institute of Science and Technology}
      \city{Moscow} 
      \country{Russia}
}

\author{Sergei Satsevich}
\email{Sergei.Satsevich@skoltech.ru}
\orcid{0009-0002-7666-8587}
\affiliation{%
      \institution{Skolkovo Institute of Science and Technology}
      \city{Moscow} 
      \country{Russia}
}

\author{Olga Shalopanova}
\email{Olga.Shalopanova@skoltech.ru}
\orcid{0009-0002-5748-1979} 
\affiliation{%
      \institution{Skolkovo Institute of Science and Technology}
      \city{Moscow} 
      \country{Russia}
}

\author{Miguel Altamirano Cabrera}
\email{miguel.altamirano@skoltech.ru}
\orcid{0000-0002-5974-9257}
\affiliation{%
      \institution{Skolkovo Institute of Science and Technology}
      \city{Moscow} 
      \country{Russia}
}

\author{Artem Lykov}
\email{artem.lykov@skoltech.ru}
\orcid{0000-0001-6119-2366} 
\affiliation{%
      \institution{Skolkovo Institute of Science and Technology}
      \city{Moscow} 
      \country{Russia}
}

\author{Dzmitry Tsetserukou}
\orcid{0000-0001-8055-5345}
\email{d.tsetserukou@skoltech.ru}
\affiliation{%
      \institution{Skolkovo Institute of Science and Technology}
      \city{Moscow} 
      \country{Russia}
}

\renewcommand{\shortauthors}{Bazhenov et al., 2024}

\begin{abstract}
This paper introduces DogSurf - a new approach of using quadruped robots to help visually impaired people navigate in real world. The presented method allows the quadruped robot to detect slippery surfaces, and to use audio and haptic feedback to inform the user when to stop. A \textbf{state-of-the-art GRU-based neural network architecture with mean accuracy of 99.925\% was proposed} for the task of multiclass surface classification for quadruped robots. A dataset was collected on a Unitree Go1 Edu robot. The dataset and code have been posted to the public domain.
\end{abstract}






\maketitle

\section{Introduction}

\begin{figure}[hbt]
    \centering
    \includegraphics[width=8cm]{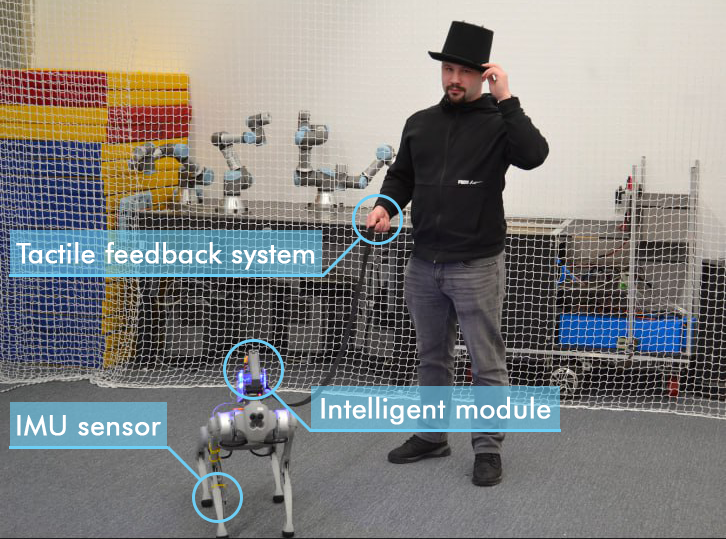}
    \caption{DogSurf system for guiding the visually impaired individuals}
    \label{fig:Key elements of DogSurf system}
\end{figure}

Recent years are marked with rapid progress of intelligent robotics. The robots are expected to step out of the laboratory and to find promising real-world applications. One of them is the use of walking robots as assistants for visually impaired people. Utilizing such assistants will help people to have more independent lives, and to spend less effort interacting with their surroundings. 

Currently, it is common to use guide dogs for this purpose, but the high cost of training and maintenance poses challenges to widespread accessibility. Robotic guide dogs, however, have the potential to be more affordable.

The initial breakthrough in using mobile robots as guide dogs was achieved in 1981 by Professor Susumu Tachi et al. \cite{tachi1981guide}. However, technological limitations at that time hindered the widespread adoption of the project. Significant advancements have emerged only four decades later.

For example, work conducted by Guerreiro et al. \cite{guerreiro2019cabot} has focused on the development of wheeled mobile platforms for assisting blind individuals in navigation. The research conducted by Xiao et. al. \cite{Xiao_Robotic_Guide_Dog} introduced a quadruped robot with a leash able to navigate through tight and congested areas. Chen et al. \cite{Chen_A_Comfort-Based_Approach} proposed a new control system for a guide robot based on an elastic rope and a thin string using a motor to actively change the length of the leash.  This allowed much better control of human movement. Hwang et al. \cite{Hwang} proposed a local path planner that considers both human and dog positions.

Overall, these advancements in robot control systems hold the promise of revolutionizing the lives of visually impaired individuals by providing them with reliable and affordable robotic assistants that can effectively navigate them through surroundings.

Operating in a real-world environment requires sufficient safety skills from walking robots. One of them is the robot capability to assess the characteristics of the surface, especially its slipperiness. A slippery surface can cause severe trauma as a result of fall, for example, hip fractures. The risk of fracture increases with age, making it especially dangerous after the age of 60 \cite{Sing_CW_bones}. Furthermore, the total number of hip fractures in 2050 is projected to be approximately twice the number reported in 2023 \cite{Sing_CW_bones}.

As a result, a robotic guide dog must be smart enough to protect the human from fall on a slippery surface.

\section{Related works}
Multiple methodologies exist for surface recognition in a robot locomotion.

\textbf{First approach} is to \textbf{use different types of tactile estimation}. Mudalige et al. \cite{Nipun} proposed to use Touch Sensitive Foot (TSF) tactile sensors. They were placed on the dog’s feet to collect data, which was later processed by the Convolutional Neural Networks (CNN). This approach allows achieving 74.37\% validation accuracy and highest recognition score of 90\% for line patterns.  Although this approach shows promise, there are multiple limitations, including high service cost because of the direct contact of sensors with the surface that causes gradual wear and tear over time.

\textbf{Second approach} is to \textbf{use audio information}. Dimiccoli et al. \cite{Dimiccoli} used a gripper to interact with the objects and recorded audio signals, which were later used to train CNN. This method resulted in about 85\% accuracy. Another good result was achieved by Vangen et al. \cite{Vangen}. They proposed to use sensorized paw with audio-based classification of terrain and achieved about 78\% accuracy. Overall, this approach shows limited applicability in real-world scenarios as they include a diversity of ambient sounds, which significantly impair the accuracy of the neural network.

\textbf{Third approach} is to \textbf{use an Inertial Measurement Unit (IMU) sensor.} For wheeled robot with an IMU sensor such approach was proposed by Lomio et al. \cite{lomio}. They collected data from accelerometer (3 axis), gyroscope (3 axis) for 9 classes of surfaces and tested several Machine Learning algorithms and Deep Learning architectures. XGBoost, Fully Connected Network (FCN), and ResNet (CNN)  algorithms achieved the accuracy of 59.54\%, 62.69\%, and 64.95\%, respectively. Singh et al. \cite{singh} propose to use another CNN architecture with an overall classification accuracy of 88\%.

\textbf{Fourth approach} is to use \textbf{Force/Torque (F/T) sensor} data alone or in combination with the methods described above. Bednarek et al. \cite{Bednarek_F/t} proposed to use data only from F/T sensor on quadrupedal robot ANYmal \cite{ANYmal}. Additionally, the authors proposed a new CNN-1d convolution architecture and developed clustering method for terrain classification, and achieved 93\% accuracy. Another approach with a Transformer architecture \cite{Transformers} was proposed in \cite{HAPTR2} and achieved 97.33\% accuracy on a QCAT dataset.
Kolvenbach et al. \cite{Kolvenbach_f/c_2} proposed to use combination of F/T and IMU sensors for ANYmal robot and developed special stand for tests. They achieved 98.6\% accuracy.

IMU-based approach gives better results on average, has no direct contact with the environment, it is weather- and time-of-day resistant, and cheap. It was applied to a walking robot using gated recurrent units (GRU) \cite{GRU}.


\section{DogSurf System Overview}
\subsection{Hardware architecture}
\begin{figure}[htp]
    \centering
    \includegraphics[width=8cm]{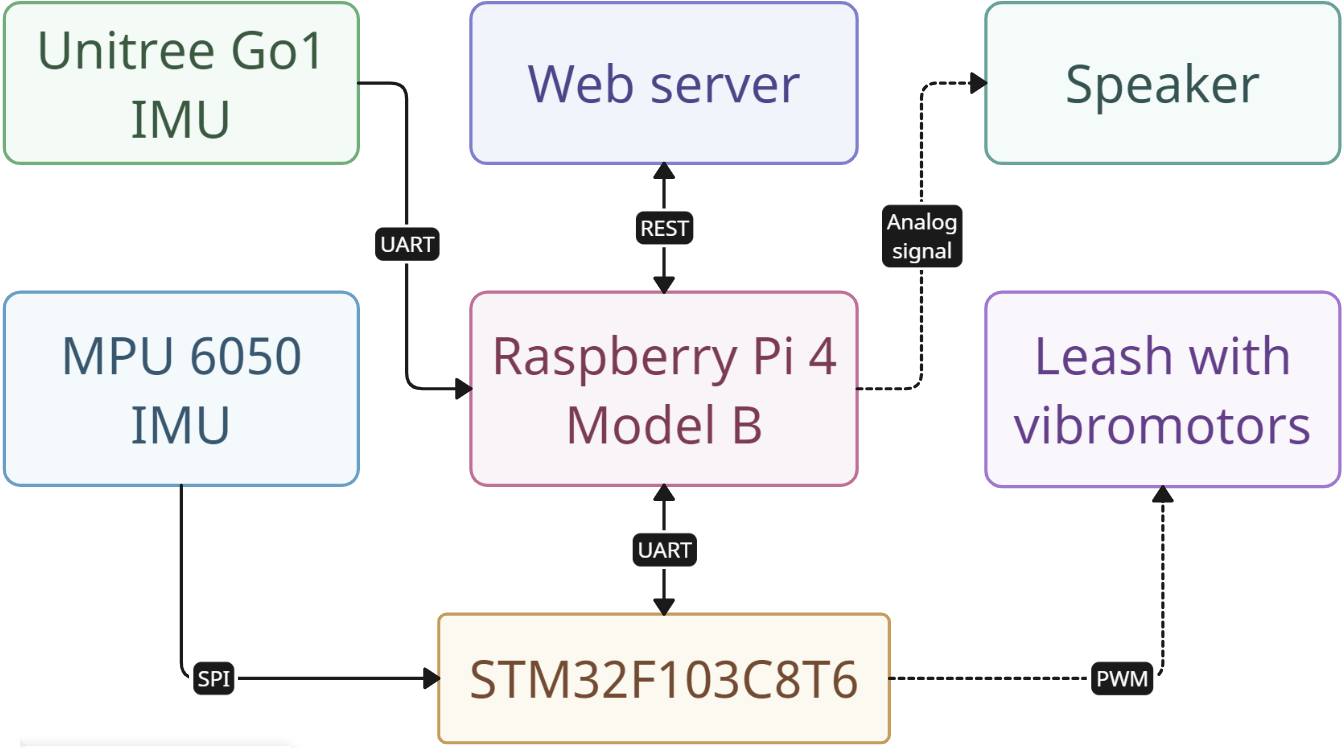}
    \caption{DogSurf system architecture}
    \label{fig:system architecture}
\end{figure}

The reserch platform for this study was Unitree Go1 Edu. The data sources used are built-in IMU sensor inside the robot and one external IMU sensor attached to the right front leg. The architecture of the Human-interaction module is presented in the (\autoref{fig:system architecture}). This module is also used for the data collection.

MPU 6050 IMU sensor \cite{mpu_6050} includes an accelerometer and gyroscope. It provides enough information for high quality surface recognition.
MCU collects data from the IMU and sends it to the Raspberry Pi to analyse the received information and to deliver a feedback through vibration motors and a speaker if the surface is slippery. Data received and logs collected are stored on a web server.

\subsection{Dataset structure}
During the research a dataset was collected with more than 900,000 samples (about 10 hours of data). Data was gathered for 5 classes of which 3 classes are wood, rubberized concrete, and tile and another 2 classes are slippery surfaces, i.e. ice and wet smooth rubberized concrete. During collecting, the robot dog was controlled from a gamepad with variation of speed and direction of movement to achieve sufficient data diversity.
 \textbf {The dataset can be accessed by the link: https://github.com/Eterwait/DogSurf}. It consists of two parts: internal data from robot dog and external data from MPU 6050. Frequency of measurements was 50 Hz. Internal data includes raw information from an accelerometer and a gyroscope, RPY (roll, pitch, yaw), and quaternion representations. External data includes only raw information from an accelerometer and a gyroscope.

\subsection{GRU-based surface recognition}
DogSurf approach utilizes a GRU neural network as the foundation. The IMU data from the accelerometer and gyroscope, is collected and appended to a queue containing 100 samples. During each iteration, a Forward Pass is conducted through the firstly \textbf{Standart Scaler}, then through \textbf{Principal component analysis (PCA) transform}, and then trough \textbf{bidirectional GRU} - neural network. This process continues until the system detects the presence of a slippery surface for three consecutive instances. Subsequently, the robot stops moving, activates vibration motors integrated into the leash and the voice alert stating: "Stop, slippery surface!"

\begin{figure}[hbt]
    \centering
    \includegraphics[width=8cm]{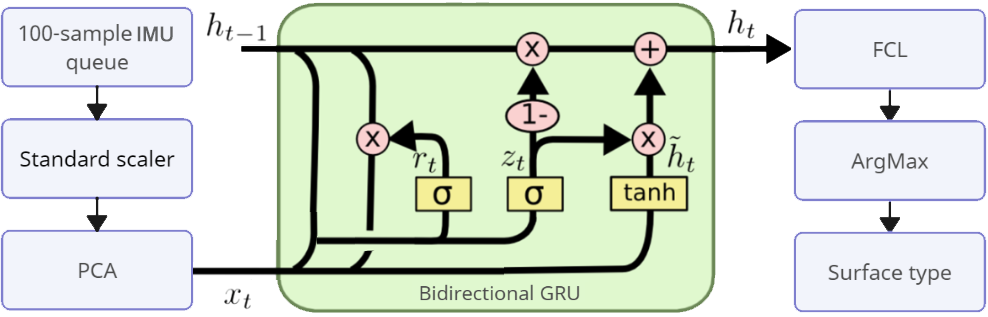}
    \caption{Neural network architecture}
    \label{fig:Neural network architecture}
\end{figure}

GRU is designed for analyzing the time series, in this case it was necessary to select the optimal number of sequential samples. A value of 100 was chosen experimentally. Data prepossessing consisted of creating a new dataset by sliding window with size of 100 and step 1 over the old dataset. After prepossessing, new dataset of  90,000,000 samples was collected. It ensures that the optimal accuracy and length of the time series are achieved. \textbf {Code can be accessed by link: https://github.com/Eterwait/DogSurf.} The work uses two data sources: the internal IMU Unitree Go1 and the external sensor MPU-6050 attached to the robot’s leg.
 The results of DogSurf approach compared to the state-of-the-art ones are listed in \autoref{tab:results}.

\begin{table}[htp]
\centering
\caption{Comparison of DogSurf with the state-of-the-art approaches}
\label{tab:results}
\begin{tabular}{c|c|c|}
  \cline{2-3}
  & Model & Accuracy\\
  \hline
   \multicolumn{1}{|c|}{Tactile} & Weerakkodi et al. \cite{Nipun} & 0.74370\\
   \hline
   \multicolumn{1}{|c|}{\multirow{2}{*}{Audio}} & Dimiccoli et al. \cite{Dimiccoli} & 0.84700\\
   \cline{2-3}
   \multicolumn{1}{|c|}{}& Vangen et al. \cite{Vangen} & 0.77900\\
   \hline
   \multicolumn{1}{|c|}{\multirow{2}{*}{F/T}} & HAPTR2 \cite{HAPTR2} & 0.97330\\
   \cline{2-3}
   \multicolumn{1}{|c|}{} & Jakub Bednarek et al.  \cite{Bednarek_F/t} & 0.93590\\
   \hline
   \multicolumn{1}{|c|}{F/T \& IMU} & Kolvenbach et al. \cite{Kolvenbach_f/c_2} & 0.98600\\
   \hline
   \multicolumn{1}{|c|}{\multirow{3}{*}{IMU}} & Lomio et al. \cite{lomio} & 0.64950\\
   \cline{2-3}
   \multicolumn{1}{|c|}{}& Singh et al. \cite{singh} & 0.88000\\
   \cline{2-3}
   \multicolumn{1}{|c|}{} & \textbf{DogSurf} & \textbf{0.99925}\\
  \hline
\end{tabular}
\end{table}

During experiments we have chosen optimal structure and number of parameters for GRU neural network. Additionally, we have compared different representations of information: raw data from accelerometer and gyroscope, RPY (roll, pitch, yaw), and quaternions. Experiments revealed that RPY transformation does not improve results, but requires additional calculations. Quaternions worsened the result, so this approach was not implemented. Comparison of internal and external data processing is shown in \autoref{tab:all results}.

\begin{table}[htp]
\centering
\caption{DogSurf results for two classes: Classical (around all 5 types of surfaces, Slippery (ice and wet smooth rubberized concrete are unified into one class)}
\label{tab:all results}
\begin{tabular}{l|c|c|c|c|}
   \cline{2-5}
  & \multicolumn{4}{|c|}{Accuracy}\\
  \cline{2-5}
  & \multicolumn{2}{|c|}{0.9k parameters} & \multicolumn{2}{|c|}{3.7k parameters}\\
  \cline{2-5}
  & Int. & Ext. & Int. & Ext.\\
  \hline
  \multicolumn{1}{|c|}{Classical} & 0.98575 & 0.99698 & 0.99065 & \textbf{0.99925}\\
  \hline
  \multicolumn{1}{|c|}{Slippery} & 0.99881 & 0.99913 & 0.99858 & 0.99992\\
  \hline
\end{tabular}
\end{table}

After fine tuning process, two types of neural networks for classical terrain classification between each type of surface, i.e. \textbf{Large} and \textbf{Small}, were proposed. \textbf{Large} one has 3.7k parameters and reduces error compare to the last State Of The Art (SOTA) algorithm in 18 times. \textbf{Small} one has 0.9k parameters resulting in a loss in accuracy of only 0.3\%.The results can be seen in \autoref{tab:results-surf}. 
But the DogSurf approach does not need to classify the different types of slippery surface, it is needed to know whether it is slippery or not with maximum probability. For this task the results can be seen in \autoref{tab:all results}. 

\begin{figure}[hbt]
    \centering
    \includegraphics[width=8cm]{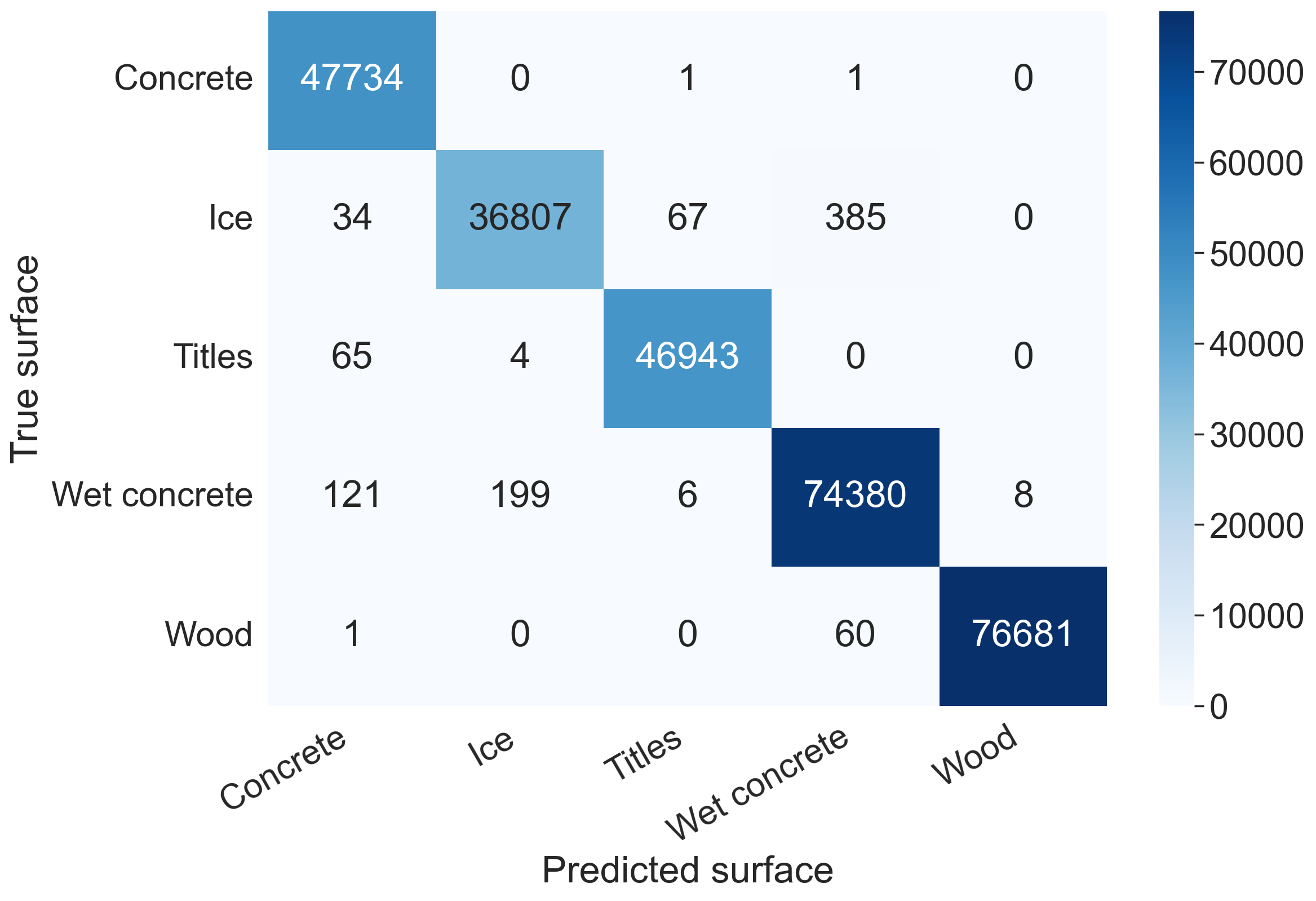}
    \caption{Confusion matrix for DogSurf with 3.7k parameters}
    \label{fig:cm}
\end{figure}

\begin{table}[htp]
\centering
\caption{Results for different surfaces for Large DogSurf}
\label{tab:results-surf}
\begin{tabular}{c|c|c|c|}
  \cline{2-4}
  & Precision & Recall & F1 score\\
  \hline
  \multicolumn{1}{|c|}{Concrete} & 0.99956 & 0.99998 & 0.99977\\
  \hline
  \multicolumn{1}{|c|}{Ice} & 0.99661 & 0.99825 & 0.99743\\
  \hline
  \multicolumn{1}{|c|}{Titles} & 0.99964 & 0.99994 & 0.99979\\
  \hline
  \multicolumn{1}{|c|}{Wet concrete} & 0.99933 & 0.99872 & 0.99903\\
  \hline
  \multicolumn{1}{|c|}{Wood} & 1.0000 & 0.99936 & 0.99968\\
  \hline
\end{tabular}
\end{table}

The outstanding results of surface recognition by DogSurf approach (see Table 3) allow to be calm about the accuracy of the model and focus in other directions of research to bring robots into the lives of the visually impaired in the near future.

\section{User Study}
A series of experiments were carried out to evaluate the user performance while using the proposed system DogSurf. Users were asked to follow the instructions of the robot by using three modalities of the system: only audio, only tactile, and the combination of audio and tactile feedback.

\subsection{Experimental procedure}
\textbf{Experiment setup:} Initially, the robot is placed at a starting point, and the users are asked to hold the rope.\newline
\textbf{Participants:} Six participants, two females and four males, capable of performing the experiment, aged 26.1$\pm$3.9 years, volunteered to participate in the experiments.\newline
The task for the subjects wearing the light blocking mask was to navigate according to the robot signals and stop when a slippery surface was detected by the system, using only audio, only tactile, and the combination of audio and tactile stimuli at the same time.

Before the experiment participants had time to familiarize themselves with the system. During the experiment, they were asked to stop walking when the signals showed that a slippery surface was detected.

\subsection{NASA-TLX results}

\begin{figure}[htp]
    \centering
    \includegraphics[width=8cm]{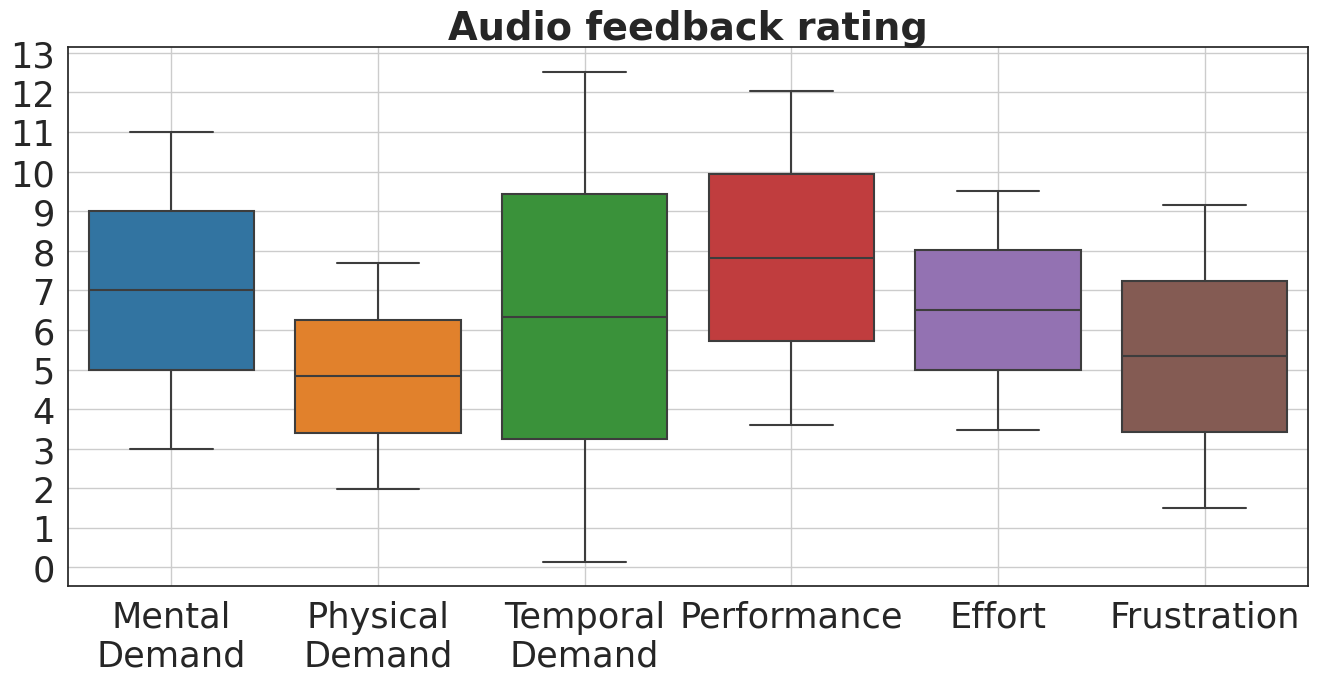}
    \caption{NASA-TLX average rating for the user navigation by DogSurf with audio feedback}
    \label{fig:NASA-TLX average rating for the robot telemanipulation1}
\end{figure}

\begin{figure}[htp]
    \centering
    \includegraphics[width=8cm]{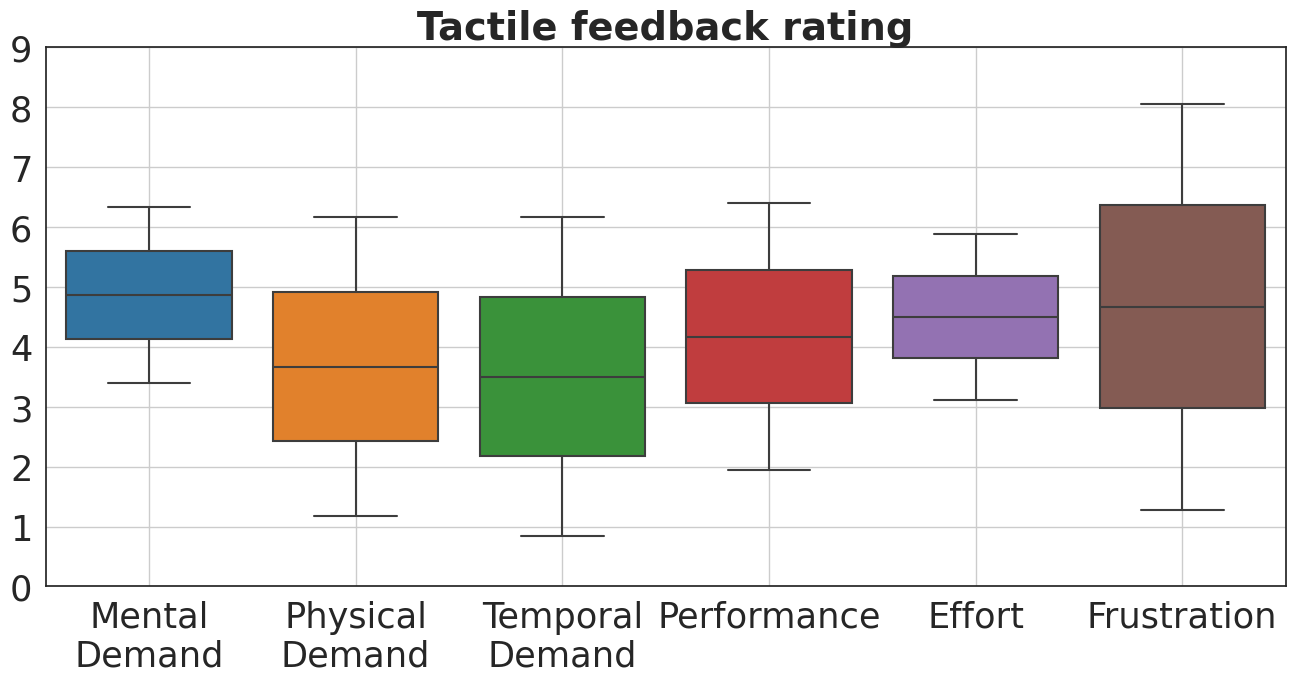}
    \caption{NASA-TLX average rating for the user navigation by DogSurf with tactile feedback}
    \label{fig:NASA-TLX average rating for the robot telemanipulation2}
\end{figure}

\begin{figure}[htp]
    \centering
    \includegraphics[width=8cm]{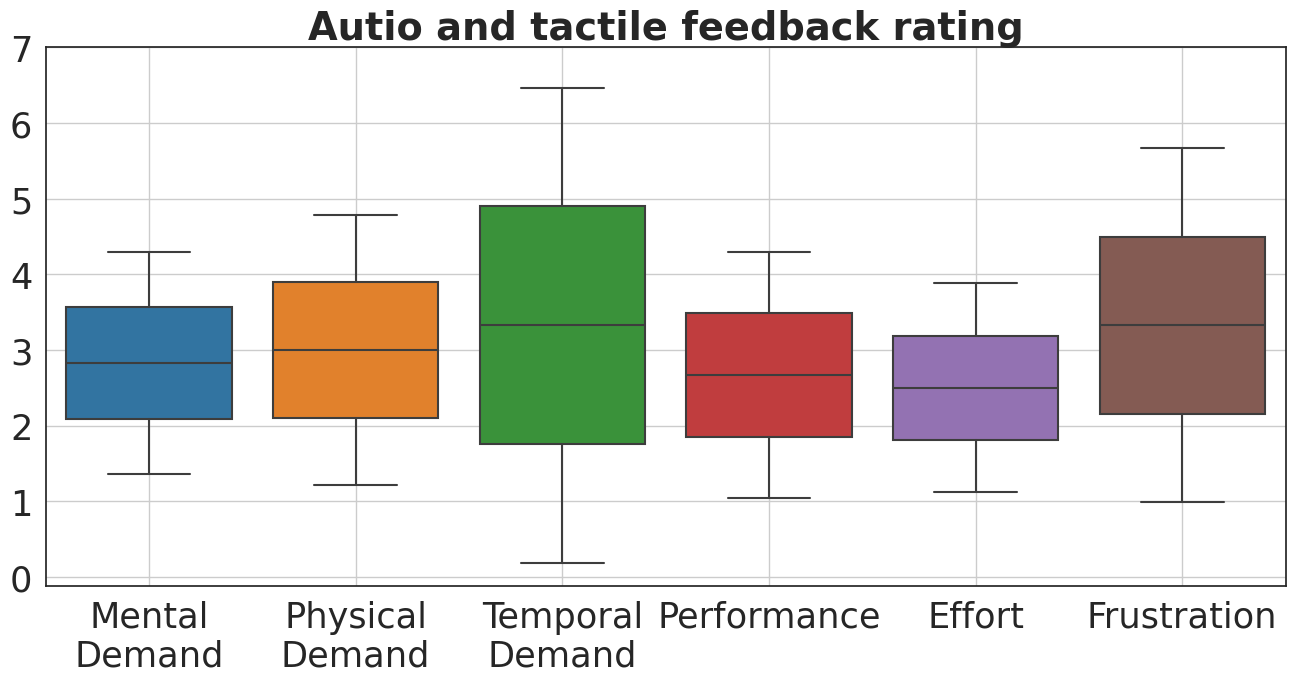}
    \caption{NASA-TLX average rating for the user navigation by DogSurf with audio \& tactile feedback}
    \label{fig:NASA-TLX average rating for the robot telemanipulation3}
\end{figure}

Friedman tests were used to analyze the results. Statistically significant differences were found in mental demand ($p=0.0421 < 0.05$), physical demand ($p=0.0223 < 0.05$), performance ($p=0.0136 < 0.05$),  effort ($p=0.0149 < 0.05$), and frustration ($p=0.0241 < 0.05$). Wilcoxon tests were performed for pairwise comparisons for the five sub-scales that presented statistical differences. According to the statistical significance difference, it was found that only  using the tactile feedback is more mental demand than using tactile plus audio ($p=0.0312$), according to the performance. There is statistical difference between only audio and audio plus tactile ($p=0.0421$). It was found that is less effort when the audio plus tactile is implemented in comparison with only audio ($p=0.0431$), and only tactile ($p=0.0421$). The frustration is decreasing when the tactile plus audio is implemented in comparison with just the audio ($p=0.0393$). The experimentsl results are summarized in \autoref{fig:NASA-TLX average rating for the robot telemanipulation1} - \autoref{fig:NASA-TLX average rating for the robot telemanipulation3}.

\section{Conclusion}
We have developed a robotic system DogSurf that allows visually impaired people to effectively avoid falling on slippery surfaces while following a robotic guide dog. Moreover, a SOTA algorithm was developed for surface recognition and especially for slippery surface detection, maximizing the accuracy and minimizing the size of the neural network among other approaches.

In future, the data from each of the robot dog's feet will be added, so that it will be possible to accurately determine whether it is standing on a slippery surface. Additional tactile patterns will be designed to navigate user in cluttered environments and on uneven surfaces (e.g. stairs). The functionality of real-time recognition of obstacles on the road, such as open manholes, traffic and construction barriers, curbs, and etc. will be provided.

\bibliographystyle{ACM-Reference-Format}
\balance
\bibliography{sample-base}
\end{document}